\documentclass[letterpaper, 10 pt, conference]{ieeeconf}  

\IEEEoverridecommandlockouts                              

\usepackage{graphicx} 
\usepackage{amsmath} 
\usepackage{amssymb}  
\usepackage{cite}
\usepackage{caption}
\usepackage{enumerate}
\usepackage{subcaption}
\usepackage{hyperref}
\usepackage[linesnumbered,ruled]{algorithm2e}
\usepackage{multirow}
\usepackage{makecell}
\usepackage{hhline}
\usepackage{amsmath}

\usepackage{subcaption}
\usepackage{color}

\title{\huge AquaMILR: Mechanical intelligence simplifies control of undulatory robots in cluttered fluid environments}

\author{Tianyu Wang$^{1,\dagger}$, Nishanth Mankame$^{1,\dagger}$, Matthew Fernandez$^{1}$, Velin Kojouharov$^{2}$, and Daniel I. Goldman$^{1}$%
\thanks{$^{1}$Tianyu Wang, Nishanth Mankame, Matthew Fernandez and Daniel I. Goldman are with Georgia Institute of Technology, Atlanta, Georgia, USA.
        {\tt\footnotesize tianyuwang@gatech.edu, nmankame3@gatech.edu, mfernandez64@gatech.edu, daniel.goldman@physics.gatech.edu}}%
\thanks{$^{2}$Velin Kojouharov is now with Stanford University, Stanford, California, USA.
        {\tt\footnotesize vkojo@stanford.edu}}
\thanks{$\dagger$These authors contributed equally to this work.}
}

\begin{document}
\maketitle

\begin{abstract}
While undulatory swimming of elongate limbless robots has been extensively studied in open hydrodynamic environments, less research has been focused on limbless locomotion in complex, cluttered aquatic environments. Motivated by the concept of mechanical intelligence~\cite{wang2023mechanical}, where controls for obstacle navigation can be offloaded to passive body mechanics in terrestrial limbless locomotion, we hypothesize that principles of mechanical intelligence can be extended to cluttered hydrodynamic regimes. To test this, we developed an untethered limbless robot capable of undulatory swimming on water surfaces, utilizing a bilateral cable-driven mechanism inspired by organismal muscle actuation morphology to achieve programmable anisotropic body compliance. We demonstrated through robophysical experiments that, similar to terrestrial locomotion, an appropriate level of body compliance can facilitate emergent swim through complex hydrodynamic environments under pure open-loop control. Moreover, we found that swimming performance depends on undulation frequency, with effective locomotion achieved only within a specific frequency range. This contrasts with highly damped terrestrial regimes, where inertial effects can often be neglected. Further, to enhance performance and address the challenges posed by nondeterministic obstacle distributions, we incorporated computational intelligence by developing a real-time body compliance tuning controller based on cable tension feedback. This controller improves the robot's robustness and overall speed in heterogeneous hydrodynamic environments.
\end{abstract}



\vspace{-0.1em}
\section{Introduction}

Biologically inspired swimming robots have emerged as compelling candidates for navigating aquatic environments~\cite{katzschmann2018exploration,zhu2019tuna,thandiackal2021emergence,baines2022multi,wang2023versatile,ren2021research}. These robots offer versatile swimming modes and even enhanced speed and efficiency over conventional underwater vehicles in their application scenarios~\cite{chu2012review,raj2016fish,wang2020development,qu2024recent,bogue2015underwater}. Specifically, elongate limbless (or snake-like, anguilliform) robots inspired by undulatory organisms, spanning from meter-scaled sea snakes and eels to millimeter-scaled nematodes, exhibit remarkable agility and maneuverability~\cite{li2023underwater,crespi2008online,thandiackal2021emergence,liljeback2014mamba,zuo2009serpentine}. This is due to their hyper-redundant body structure and ability to generate full-body undulation. However, despite extensive research in homogeneous hydrodynamic regimes, their locomotion in heterogeneous aquatic environments, particularly in highly cluttered scenarios where the body consistently interacts with multiple obstacles, remains less systematically studied.

Undulatory locomotion of limbless robots in homogeneous hydrodynamic environments has been studied for decades. Extensive research has demonstrated the efficacy and energy efficiency of this form of locomotion~\cite{li2023underwater,yamada2005development,crespi2008online,liljeback2014mamba}. By elucidating the principles underlying propulsion mechanisms, body morphology, and control strategies in controlled settings, these studies have significantly advanced our understanding of robotic undulatory swimming. However, in practical applications where challenges posed by environmental heterogeneities cannot be neglected, developing mechanisms to deal with obstacles becomes essential. Most existing systems employ perception-based obstacle-avoiding methods to prevent body-obstacle interactions~\cite{sanfilippo2017perception,kelasidi2019path}. However, these approaches require large onboard computational power and/or high-quality communication and data transmission for off-board processing. More importantly, they become less effective in obstacle-rich regions where multiple interactions occur simultaneously along the body and cannot be avoided.

\begin{figure}[t]
\centering
\includegraphics[width=0.82\columnwidth]{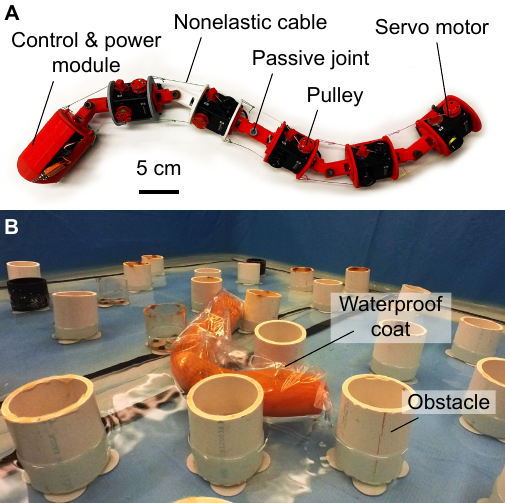}
\caption{The untethered mechanically intelligent limbless robot AquaMILR for undulatory locomotion in cluttered fluid environments. (A) The robot implements a decentralized bilateral actuation mechanism. Inset shows the design of the head module integrating power, computation, and communication modalities. (B) The robot navigates a laboratory model of an obstacle-rich environment.}
\vspace{-1.8em}
\label{fig:robot}
\end{figure}

Although not widely applied to hydrodynamic scenarios, obstacle adaptation (or obstacle-aided)~\cite{transeth2008snake,chong2023gait} mechanisms offer another major approach for limbless robot locomotion in obstacle-rich terradynamic regimes, apart from the obstacle-avoiding approach. Unlike methods that rely heavily on sensing and environmental knowledge~\cite{wang2020directional,ramesh2022sensnake}, methods that realize body compliance through purely passive mechanics~\cite{wang2023mechanical,fu2020robotic,boyle2012adaptive,liu2023nonbiomorphic} are particularly suited for aquatic applications, as they minimize the computation and communication required. Specifically, \cite{wang2023mechanical} developed a robot (Mechanically Intelligent Limbless Robot, MILR) that models the musculoskeletal actuation mechanisms of snakes and nematodes. By comparing biological and robotic locomotion kinematics, this work revealed principles of mechanical intelligence in terrestrial locomotion, demonstrating that controls for obstacle navigation can be offloaded to passive body mechanics, and mechanical intelligence alone is sufficient to handle a high density of obstacles without the need for active feedback controls.

\begin{figure*}[t]
\centering
\includegraphics[width=0.82\textwidth]{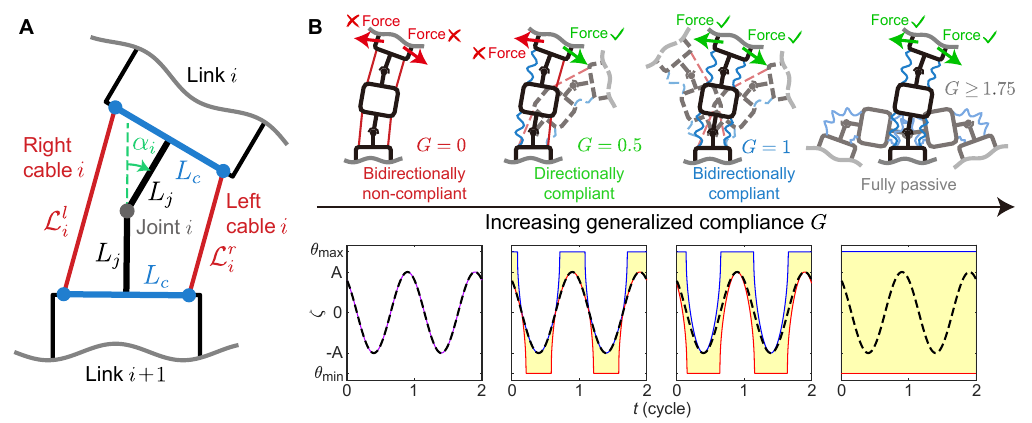}
\caption{Programmable and quantifiable body compliance through bilateral cable actuation mechanism. (A) A geometric model illustrating a single joint, used to determine the exact lengths of the left and right cables ($\mathcal{L}^l$ and $\mathcal{L}^r$) necessary to achieve a specified joint angle ($\alpha$). (B) A schematic displaying various compliance states based on the generalized compliance variable $G$: bidirectionally non-compliant ($G = 0$), where the joint remains rigid and strictly follows the template trajectory (dashed line); directionally compliant ($G = 0.5$), where the joint allows movement that increases the angle (shown by the yellow region limited by upper and lower bounds indicated by blue and red lines); bidirectionally compliant ($G = 1$), where the joint permits movement in both directions with different degrees of flexibility; and fully passive ($G \geq 1.75$), where the joint can move freely in either direction. Figures adapted from~\cite{wang2023mechanical}.}
\vspace{-1.8em}
\label{fig:explain_G}
\end{figure*}

In this work, we extend the bilateral actuation mechanism and the principles of mechanical intelligence to the hydrodynamic regime. Unlike terrestrial environments where friction dominates locomotion and inertial effects are usually negligible, hydrodynamics introduces complexities associated with fluid dynamics, buoyancy, and drag. In aquatic environments at intermediate Reynolds numbers, inertial effects become significant~\cite{carling1998self,gazzola2015gait}; the robot cannot stop moving immediately upon stopping self-deformation, leading to a coasting behavior in the fluid~\cite{justus2024optimal,rieser2024geometric}. Such behavior can profoundly influence the dynamics of undulatory propulsion, especially in heterogeneous environments. Therefore, while mechanical intelligence holds promise for simplifying control mechanisms, its application in hydrodynamic settings necessitates a systematic robophysical~\cite{aguilar2016review} study of how varying levels of mechanical intelligence influence locomotive performance. By unraveling these complexities, we aim to develop novel strategies for navigating cluttered aquatic environments with undulatory limbless robots, ultimately advancing undulatory robotic locomotion in inertial fluid mediums.

\vspace{-0.3em}
\section{Untethered Robot Development and Control}

\subsection{Robot design and manufacturing}
To investigate principles of mechanical intelligence in heterogeneous aquatic environments, we developed a modular limbless robot based on the MILR design~\cite{wang2023mechanical}. Our 66-cm-long hard-soft hybrid model features 5 bending joints (Fig.~\ref{fig:robot}A) and employs a bilateral actuation mechanism. Passive bending joints are controlled by adjusting cable lengths via decentralized cable-pulley-motor systems, with each cable independently managed on either side of the joint. By synchronizing the cable lengths through sequential angular oscillations along the body, the robot generates undulatory motion.

In intermediate Reynolds numbers aquatic environments, locomotion is highly sensitive to inertial effects, where even small forces can disrupt overall performance. To minimize such perturbations, we employed an untethered design. Specifically, we designed a control and power module housed in the robot's head (Fig.~\ref{fig:robot}A). This head module integrates a single-board computer, a motor communication converter, a voltage regulator, and a battery.

The entire robot body, including the casing, joint links, and pulleys, was 3D printed using PLA materials. For the cable-pulley-motor systems, we controlled the lengthening and shortening of the nonelastic cables (Rikimaru Zero Stretch \& Low Memory) with Dynamixel 2XL430-W250-T (Robotis) servo motors, which provide up to 1.4 Nm torque. To waterproof the system, a tube-shaped polyethylene plastic sleeve was used to encase the robot (Fig.~\ref{fig:robot}B). The robot was placed inside a section of tubing, and the plastic was heat-sealed at both ends to create a watertight seal. This design allowed the robot to effectively swim on the surface of the water.

The head module featured a removable top and housed a single-board computer (Raspberry Pi Zero 2W) for onboard computation, real-time communication with a PC via Wi-Fi, and real-time control of the servo motors through a Dynamixel U2D2. A 1000-mAh 3-cell LiPo battery powered all the motors and the single-board computer: the motors received 12V, while a 5V/2.5A voltage regulator stepped down the voltage for the computer.

We named the robot, following MILR as in~\cite{wang2023mechanical}, as AquaMILR, and it will be referenced as such throughout the remainder of this paper.

\vspace{-0.3em}
\subsection{Suggested gait}
To achieve a basic traveling-wave locomotion pattern on AquaMILR, we designed a shape control scheme based on the ``serpenoid" shape template introduced by~\cite{hirose1993biologically}. This template allows a wave to move from the head to the tail of the robot when the angle of the $i$-th joint angle, $\alpha_i$, at time $t$ is given by
\vspace{-0.7em}
\begin{equation}
\begin{aligned}
\alpha_i(t) = A\sin(2\pi\xi \frac{i}{N} - 2\pi\omega t),
\label{eq:template}
\end{aligned}
\end{equation}
where $A$ denotes the amplitude, $\xi$ is the spatial frequency, $\omega$ is the temporal frequency, $i$ is the joint index, and $N$ is the total number of joints. This angle $\alpha_i$ is referred to as ``suggested" angle. 

To precisely achieve the joint angle $\alpha$ as defined in Eq.~\ref{eq:template}, it is necessary to adjust the lengths of the left and right cables around the joint, ensuring both are appropriately shortened (as shown in Fig.~\ref{fig:explain_G}A). Note that we use nonelastic cables, their deformation is negligible. The lengths of the left cable ($\mathcal{L}^l$) and right cable ($\mathcal{L}^r$) are determined by the robot's geometry then, as illustrated in Fig.~\ref{fig:explain_G}A, and follow these equations:
\vspace{-0.7em}
\begin{equation}
\begin{aligned}
    \mathcal{L}^l(\alpha) &= 2\sqrt{L_{c}^2 + L_{j}^2} \cos\left[-\frac{\alpha}{2}+\tan^{-1}\left(\frac{L_{c}}{L_{j}}\right)\right],\\
    \mathcal{L}^r(\alpha) &= 2\sqrt{L_{c}^2 + L_{j}^2} \cos\left[\frac{\alpha}{2}+\tan^{-1}\left(\frac{L_{c}}{L_{j}}\right)\right].
\end{aligned}
\label{eq:ExactLength}
\end{equation}

\subsection{Programmable body compliance}
Using Eq.~\ref{eq:ExactLength}, we can accurately implement body postures for lateral undulation gaits in the robot. Further, the advantage of implementing the bilateral actuation mechanism is that it allows us to program body compliance by coordinately loosening the cables. By applying the generalized compliance variable ($G$) as defined in~\cite{wang2023mechanical}, we can quantify AquaMILR's body compliance. The cable length control scheme sets the lengths of each pair of left and right cables ($L_i^l$ and $L_i^r$) based on the corresponding suggested angle ($\alpha_i$):
\vspace{-0.5em}
\begin{equation}
\begin{array}{l}
L_{i}^l(\alpha_{i}) = \left\{\begin{array}{llc}{\mathcal{L}_{i}^l(\alpha_{i}),} & {\text{if } \alpha_{i} \leq -\gamma} \\ {\mathcal{L}_{i}^l[-\min(A, \gamma)]+l_0\cdot[\gamma + \alpha_{i}],} & {\text{if } \alpha_{i} > -\gamma}\end{array}\right. \\ 
L_{i}^r(\alpha_{i}) = \left\{\begin{array}{llc}{\mathcal{L}_{i}^r(\alpha_{i}),} & {\text{if } \alpha_{i} \geq \gamma} \\ 
 {\mathcal{L}_{i}^r[\min(A, \gamma)]+l_0\cdot[\gamma - \alpha_{i}],} & {\text{if } \alpha_{i} < \gamma}\end{array}\right.
\end{array}
\label{eq:policy}
\end{equation}
where the superscripts $l$ and $r$ denote left and right, respectively, and $\gamma=(2G_i-1)A$. The design parameter $l_0$ determines the relaxed length of a cable, which is fixed at 0.73 mm/degree for this work (for a full discussion on choosing $l_0$ and its derivation, refer to \cite{wang2023mechanical}). Following Eq.~\ref{eq:policy}, the robot can achieve three representative compliance states with varying $G$ (Fig.~\ref{fig:explain_G}B): 1) bidirectionally non-compliant ($G=0$), where each joint angle strictly follows the suggested gait template in Eq.~\ref{eq:template}; 2) directionally compliant ($G=0.5$), where the joints are only allowed to deviate to form a larger angle than suggested; and 3) bidirectionally compliant ($G=1$), where the joints can deviate in both directions, regulated by Eq.~\ref{eq:policy}. Note that $G\in [0,\infty)$ is a continuous value: for any value of $G$ between the representative compliance states, the joint shows a hybrid compliance state based on the real-time suggested angle $\alpha$; when $G$ exceeds 1 and keeps increasing, the joint remains bidirectional compliant and becomes more slack until it reaches a fully passive state ($G=1.75$ in this work).

\vspace{-0.7em}
\section{Experiment}
\subsection{Experiment setup}

\begin{figure}[t]
\centering
\includegraphics[width=0.82\columnwidth]{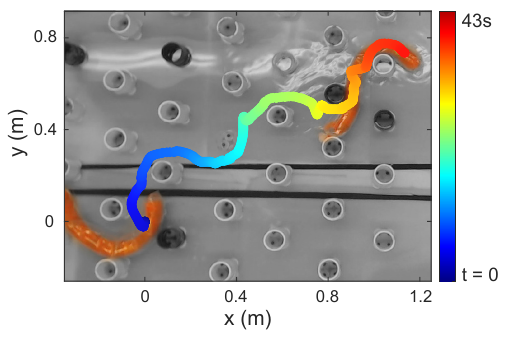}
\caption{An example of robot locomotion in a regularly distributed lattice, showing frames of the robot's starting and ending poses, along with its tracked trajectory over time.}
\vspace{-1.8em}
\label{fig:traj_track}
\end{figure}

To create a laboratory aquatic environment for testing AquaMILR's performance, we set up an indoor pool (Bestway) with a water level of 20 cm. Obstacles were constructed using 9 cm diameter PVC pipes, each equipped with suction cup feet at the bottom (Fig.~\ref{fig:robot}B). These obstacles were arranged in a triangular grid pattern with 25 cm spacing (Fig.~\ref{fig:traj_track}). In short, the triangular obstacle terrain will be referred to as the lattice throughout the rest of paper. The lattice measures 2 m by 2 m. The suction cups ensured secure attachment of the obstacles to a plexiglass sheet, providing a stable and smooth surface that is submerged to the pool floor. Consequently, the obstacles in the heterogeneous aquatic terrain utilized throughout this study maintained their shape and position without deformation or displacement upon collisions with AquaMILR.

\vspace{-0.7em}
\subsection{Robophysical experiment}
We conducted robophysical experiments to investigate how different parameters influenced the robot's capability to navigate the lattice. First we studied the effect of body compliance. In this set of experiments, we maintained a consistent $G$ value across all joints and varied it to observe performance differences. In terradynamic regimes, previous work experimentally determined that a $G$ value between 0.75 and 1 provided appropriate compliance for maneuvering through multiple types of heterogeneous terrains~\cite{wang2023mechanical}. To determine if this principle remained applicable in aquatic settings, we conducted experiments with $G$ values ranging from 0 to 1.5 in increments of 0.25.

The subsequent experiment aimed to explore the effects of gait parameter spatial frequency and amplitude, as defined in Eq.~\ref{eq:template}, on locomotion. Spatial frequency tests examined values ranging from 0.3 to 1.2 with increments of 0.3. Values exceeding 1.2 were omitted due to the robot's limited length, and the amplitude was maintained at a constant 55 degrees. For gait amplitude tests, $A=$ 40, 55, and 70 degrees were evaluated with a consistent spatial frequency of 0.6.

Further, unlike terradynamic environments where quasi-static motion and negligible inertial effects can be assumed, the influence of undulation frequency on locomotion cannot be ignored in aquatic settings. Hence, we conducted experiments to vary the temporal frequency gait parameter. Referring to the gait template Eq.~\ref{eq:template}, temporal frequency $\omega$ regulates how fast the robot undulates. The experiments tested temporal frequencies from 0.025 Hz to 0.15 Hz in increments of 0.025 Hz. For each temporal frequency value, the robot executed a gait with $G = 1$, $A=55$ and $\xi=0.6$. Frequencies exceeding 0.15 Hz were excluded as the motors were unable to achieve the desired maximum body curvature at such undulation speeds.

\subsection{Experiment protocol}
In all robophysical experiments, at the beginning of a trial, the robot was placed into the lattice with a randomly selected position and orientation. A trial concluded when the robot either became jammed within the lattice or successfully traversed it. In the results reported in the following section, distance traveled was computed as the Euclidean distance between the robot's starting and ending positions; average speed was computed as the distance traveled divided by the time taken. The trajectories of the robot's body were acquired through overhead video tracking. This tracking process involved affixing a bright dot on the robot's head, and the dot's trajectory was then extracted from videos utilizing the Adobe After Effects motion tracker plugin (Fig.~\ref{fig:traj_track}).

\begin{figure}[t]
\centering
\includegraphics[width=0.82\columnwidth]{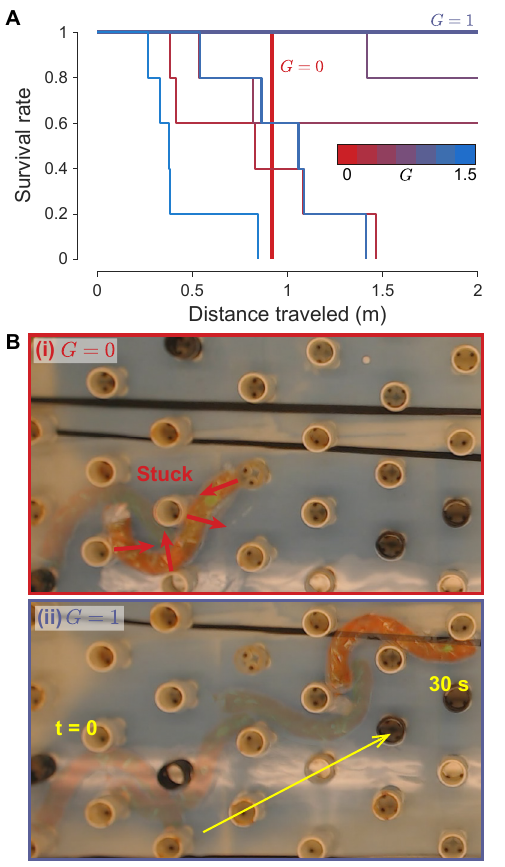}
\caption{The effect of the generalized compliance parameter ($G$) on locomotion performance. (A) The survivor function for varied $G$ values with respect to distance traveled, indicating the robot’s displacement before encountering an sticking pose or a motor failure. (B) Time-lapsed frames showing (i) an example of the robot becoming stuck at $G=0$ and (ii) an example of the robot successfully traversing the lattice at $G=1$.}
\vspace{-1.8em}
\label{fig:G_sweep}
\end{figure}
\vspace{-0.3em}
\section{Results}
\vspace{-0.2em}

\subsection{Robophysical experiments}
In this section, we present the results of our robophysical experiments. By varying the parameters of the generalized compliance ($G$), gait amplitude ($A$), spatial frequency ($\xi$), and temporal frequency ($\omega$), we conducted repeated trials with AquaMILR operating under purely feedforward, open-loop controls -- the robot generated prescribed undulation motion without having any sensing-based active adjustments to the pattern. These experiments allowed us to investigate how these parameters affect AquaMILR's behavior and performance within the lattice, discover when the mechanical intelligence emerges, and quantify the conditions that maximize it.

\subsubsection{Generalized compliance $G$}

Previous work~\cite{wang2023mechanical} demonstrated that, in the terradynamic regime, the robot exhibited the best capability to navigate through lattices within a mid-range of generalized compliance ($G$): a small $G$ frequently results in jamming poses between obstacles and a large $G$ leads to insufficient thrusting forces for moving forward. To verify whether this principle extends to the hydrodynamic regime, we experimented with the robot under varied $G$ values, using a fixed gait template that resulted in the same body wavelength-to-post spacing ratio as in the terrestrial case ($A=55, \xi=0.6, \omega=0.05$). Fig.~\ref{fig:G_sweep}A shows the survival rate as a function of the distance traveled by the robot. The results indicate that a mid-level $G$ remains optimal for navigating obstacles, as reflected by the highest survival rates over distance traveled. Specifically, $G=1$ emerged as the most appropriate value, enabling the robot to traverse the lattice (Fig.~\ref{fig:G_sweep}B-ii) in all trials. Our experiments showed that with a low level of $G<0.75$, the robot body was too rigid, and when combined with coasting dynamics in the lattice, it often ended up in jamming configurations. At a high level of $G>1$, the robot became too compliant, and the inability to maintain the desired body curvature significantly hampered propulsion through the lattice.

\begin{figure}[t]
\centering
\includegraphics[width=0.82\columnwidth]{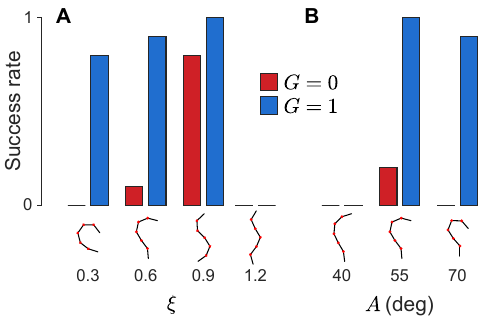}
\caption{The effect of gait parameters on locomotion performance. (A) Success traversal rate as a function of spatial frequency ($\xi$). (B) Success traversal rate as a function of amplitude ($A$).}
\vspace{-1.8em}
\label{fig:para_sweep}
\end{figure}

\begin{figure}[t]
\centering
\vspace{0.7em}
\includegraphics[width=0.82\columnwidth]{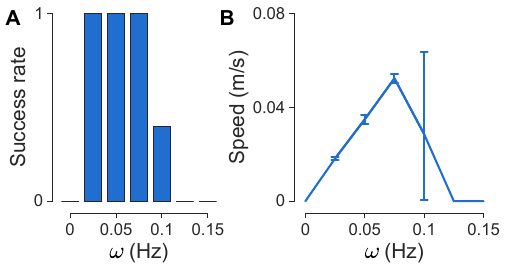}
\caption{The effect of undulation temporal frequency ($\omega$) on locomotion performance. (A) Success rate as a function of temporal frequency. (B) Averaged absolute speed of the robot as a function of temporal frequency. Error bars indicate standard deviations.}
\vspace{-1.8em}
\label{fig:tmfreq_sweep}
\end{figure}

\subsubsection{Gait spatial frequency and amplitude}
By varying gait parameters in the template~\ref{eq:template}, we investigated the robustness of the mechanical intelligence over a wide range of gaits. In this set of experiments, we tested the robot with $G = 0$ and $G=1$, the appropriate value that induced mechanical intelligence as found from the previous section. 

First, we fixed $A=55$ deg and varied the spatial frequency. Fig.~\ref{fig:para_sweep}A depicts the successful traverse rate as a function of spatial frequency for both the noncompliant robot ($G=0$) and the mechanically intelligent robot ($G=1$). The results show that $G=1$ allows the robot to traverse the lattice over a wider range of spatial frequencies than $G=0$. Also note that the robot was unable to navigate through spatial frequency values above 0.9, even with $G=1$. This is likely due to the insufficient curvature in the wave shape, which hinders the robot's ability to latch around obstacles and propel itself forward.

We then fixed $\xi=0.9$ and varied the amplitude. Similar to the results above, Fig.~\ref{fig:para_sweep}B reveals that $G=1$ allows the robot to traverse the lattice over a wider range of amplitudes than $G=0$. Overall, although for a specific lattice there is a combination of gait parameters that allows the non-compliant robot to traverse (in this case, $A=55, \xi=0.9$), mechanical intelligence reduces the sensitivity of robot performance to parameter selection, allowing a wider range of gaits to be effective.

\begin{figure*}[t]
\centering
\includegraphics[width=0.8\textwidth]{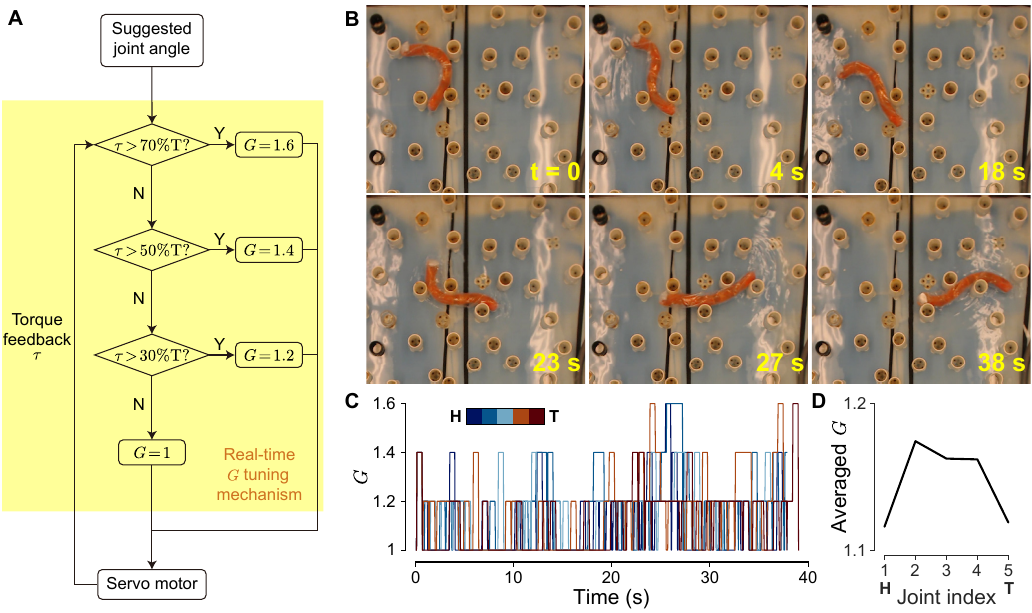}
\caption{Computational intelligence enhances mechanical intelligence, improving locomotion capabilities. (A) Block diagram of the decentralized feedback controller implementing real-time local $G$ tuning in each joint. (B) Video snapshots illustrating the robot successfully navigating a challenging, randomly distributed lattice. (C) Time records of $G$ values in individual joints from head (H) to tail (T) in the example. (D) Averaged $G$ values over time in each joint from head (H) to tail (T).}
\vspace{-1.8em}
\label{fig:G_controller}
\end{figure*}
\subsubsection{Undulation frequency}
Previous sections have verified that findings about mechanical intelligence in terrestrial environments can be extended to aquatic environments. However, the largest difference in locomotion between terradynamic and hydrodynamic regimes is the effect of inertia. In the terradynamic regime, locomotion can be assumed quasi-static, making performance insensitive to undulation frequency. Conversely, in the hydrodynamic regime, the coasting effect influences performance with increased undulation frequency leading to increased coasting. Thus, to study the effect of undulation frequency on performance, we varied the temporal frequency ($\omega$) in the gait template Eq.~\ref{eq:template}, while keeping other parameters fixed ($A=55$, $\xi=0.6$, $G=1$). Fig.~\ref{fig:tmfreq_sweep} showcases the results for the successful traverse rate and the absolute speed as functions of temporal frequency. The results first showed a linear relationship in speed from 0.025 Hz to 0.075 Hz, peaking at 0.062 m/s. When $\omega>0.075$, jamming events started to emerge at $\omega=0.1$, resulting in a large standard deviation in speed, and became dominant at higher $\omega$. Higher undulation frequency induced more unpredictable collisions, leading to abrupt deviations in the robot's trajectory (as shown in Fig.~\ref{fig:traj_track}). These collisions increased the probability of jamming instances in the lattice. This result suggests that, in hydrodynamic environments, undulation frequency plays an important role in mechanical intelligence. Relying solely on mechanical intelligence is not sufficient for a robot to operate at high frequency. Thus, to further improve absolute speed of swimming in cluttered fluid environments, computational intelligence needs to be leveraged on top of mechanical intelligence.

\subsection{Decentralized body compliance controller development}

To enhance robot performance at higher undulation frequencies and in randomly distributed lattices, which better model irregularities often found in natural environments, we developed a feedback controller that can dynamically modulate $G$ in real time based on the torque experienced in the robot joints.

\subsubsection{Controller design}
Instead of using a constant $G$ value for all joints, this controller works in a decentralized manner, i.e., modifies each joint's $G$ locally. The controller modulates the $G$ value of the $i$-th joint following
\vspace{-0.5em}
\begin{equation}
    G_i(\tau) = 1 + 0.2\sigma(\tau_i - 0.3T) + 0.2\sigma(\tau_i - 0.5T) + 0.2\sigma(\tau_i - 0.7T),
\end{equation}
where $\tau_i$ is the estimated torque experienced in the servo motors that control either the left or the right cable in the $i$-th joint, and $T$ is the maximum stall torque of the servo motor (1.4 Nm in our case). $\sigma$ represents the step function
\vspace{-0.7em}
\begin{equation}
    \sigma(x) = \begin{cases} 
      0 & x < 0 \\
      1 & x \geq 0 \\ 
   \end{cases}
\end{equation}
which realizes a three-step torque thresholding mechanism (Fig.~\ref{fig:G_controller}A). Starting from the most appropriate value $G = 1$, this mechanism allows local $G$ to increase in increments of 0.2. Further, once an increase takes place, the increased $G$ value will be maintained for 0.5 seconds or until the sensed torque drops below the thresholds.

\subsubsection{Performance test}
To test if the controller improved performance, we built a randomly distributed lattice by perturbing the regularly distributed lattice (Fig.~\ref{fig:G_controller}B). In each experiment, the robot was placed randomly in the lattice with a random orientation, and operated with $\omega=0.15$ Hz. The controller enabled the robot to achieve a 100\% successful traverse rate in the random lattice. Fig.~\ref{fig:G_controller}B, C and D show an example of a traverse, the $G$ values of all joints and the averaged $G$ values of all joints throughout the entire course, respectively. This example demonstrated that, in scenarios where mechanical intelligence alone would lead to jamming (such as $t=23$ to $27$ s), the controller adjusts the body compliance, allowing the robot to avoid jamming and continue moving. Thus, this controller shows that for challenging environments where mechanical intelligence alone is not sufficient, computational intelligence complements it and further enhances the robot's locomotion capability.

\subsubsection{Discussion}
In addition to enhancing performance, we observed during experiments that this control mechanism allowed the robot to function as a sensing tool to probe the density of its surrounding environment. For example, from $t=23$ to $27$ s, where we observed an emergent increase in $G$, indicating an increasing obstacle density around the robot. We can also roughly estimate which portion of the body experiences peak environmental forces, due to the decentralized nature of the controller. Studies have shown the importance of not only navigating unknown environments but also having the ability to characterize surroundings \cite{qian2019rapid,bush2023robotic}. Knowledge of the environment can lead to better-informed decisions for locomotion strategies on hazardous or challenging terrain. Thus, we hypothesize that, when coupled with an on-board localization and mapping system, AquaMILR with the real-time $G$ tuning mechanism can be used to probe heterogeneous aquatic environments while locomoting within them.

Video clips of robot experiments can be found in SI movie: \url{https://youtu.be/K7s5Wt3Qs14}.

\vspace{-0.5em}
\section{Conclusion}
\vspace{-0.2em}
In this work, we developed an untethered limbless robot, AquaMILR, that extends the bilateral actuation mechanism and the principles of limbless mechanical intelligence to the hydrodynamic regime. Through a series of robophysical experiments, we verified that while the general principles hold, mechanical intelligence becomes limited by the undulation frequency due to the nature of hydrodynamic environments, where inertial effects influence motion more than in terradynamic environments. Further, we developed a decentralized feedback controller that can modulate local body compliance in real time, allowing the robot to navigate through disordered heterogeneous aquatic environments at high undulation frequencies. Our results demonstrate that computational intelligence can complement mechanical intelligence, improving the robustness of locomotion.

This work paves the way for the future development of bilaterally actuated limbless robots for swimming in heterogeneous environments. Future work includes the development of onboard sensing modalities for sophisticated controller design, upgrades to enable the robot to swim at various depths and access 3D heterogeneous environments freely, and the development of amphibious robots that can transition between multiple locomotion modes, such as lateral undulation and sidewinding~\cite{kojouharov2024anisotropic}, on/in varied substrates. Further, based on this case study, we aim to further understand and model the synergy between mechanical intelligence and computational intelligence in hydrodynamic settings, which could potentially improve the locomotion efficacy and efficiency of underwater limbless robots.

\vspace{-0.5em}
\section{Acknowledgement}
\vspace{-0.3em}
The authors would like to thank Bangyuan Liu and Frank L. Hammond III for providing preliminary test environment and apparatus; Anushka Bhumkar for helping with robot maintenance and experiments; Christopher J. Pierce for useful discussions. This study is supported by Army Research Office grant (W911NF-11-1-0514) and NSF Physics of Living Systems Student Research Network (GR10003305).

\bibliographystyle{IEEEtran}

\bibliography{main_bib}

\end{document}